\documentclass[2p,12pt]{elsarticle}
\usepackage[linesnumbered,ruled,vlined]{algorithm2e}
\usepackage{hyperref}
\usepackage[numbers]{natbib}
\usepackage{adjustbox}
\usepackage{graphicx}
\usepackage{booktabs}
\usepackage{xcolor}
\usepackage{lipsum}
\usepackage{amsmath}
\usepackage{amssymb}
\usepackage{array}
\usepackage{soul}
\makeatletter
\def\ps@pprintTitle{%
  \let\@oddhead\@empty
  \let\@evenhead\@empty
  \def\@oddfoot{\reset@font\hfil\thepage\hfil}
  \let\@evenfoot\@oddfoot
}
\makeatother

\begin{document}
\definecolor{yellow}{rgb}{1,1,0} % Define yellow
\sethlcolor{yellow} % Set the highlighting color to yellow

\begin{frontmatter}


\title{Manufacturing Service Capability Prediction with Graph Neural Networks}

\author[inst1]{Yunqing Li}

\affiliation[inst1]{organization={Edward. P. Fitts Department of Industrial and Systems Engineering, North Carolina State University},%Department and Organization
            %addressline={Address One}, 
            city={Raleigh},
            postcode={27695}, 
            state={NC},
            country={USA}}
\author[inst2]{Xiaorui Liu}
\author[inst3]{Binil Starly}
\affiliation[inst2]{organization={The Department of Computer Science, North Carolina State University},%Department and Organization
            %addressline={Address One}, 
            city={Raleigh},
            postcode={27695}, 
            state={NC},
            country={USA}}

\affiliation[inst3]{organization={School of Manufacturing Systems and Networks, Arizona State University},%Department and Organization
            %addressline={Address Two}, 
            city={Mesa},
            postcode={85212}, 
            state={AZ},
            country={USA}}

\begin{abstract}
In the current landscape, the predominant methods for identifying manufacturing capabilities from manufacturers rely heavily on keyword matching and semantic matching. However, these methods often fall short by either overlooking valuable hidden information or misinterpreting critical data. Consequently, such approaches result in an incomplete identification of manufacturers' capabilities. This underscores the pressing need for data-driven solutions to enhance the accuracy and completeness of manufacturing capability identification. To address the need, this study proposes a Graph Neural Network-based method for manufacturing service capability identification over a knowledge graph. To enhance the identification performance, this work introduces a novel approach that involves aggregating information from the graph nodes' neighborhoods as well as oversampling the graph data, which can be effectively applied across a wide range of practical scenarios. Evaluations conducted on a Manufacturing Service Knowledge Graph and subsequent ablation studies demonstrate the efficacy and robustness of the proposed approach. This study not only contributes a innovative method for inferring manufacturing service capabilities but also significantly augments the quality of Manufacturing Service Knowledge Graphs.

\end{abstract}


\end{frontmatter}


\section{Introduction}
\label{sec:intro}
\subsection{Background and Motivation}
Recent global crises, including the pandemic, the shift towards near-shoring in manufacturing, and escalating geopolitical tensions, have significantly impacted supply chains across all major industries \citep{cai2020influence}. Small manufacturing companies, which are the backbone of most manufacturing economies~\cite{OFFODILE2002147}, have been disproportionately affected by supply chain disruptions, with delays in manufacturing and shipping, and shortages across the board \cite{doi:10.1080/00207543.2022.2118889}. The crises prevent these enterprises, which typically rely on interpersonal connections and regional web directories to look for new business prospects. Therefore, it is vital to implement advanced and effective tactics for the identification of small manufacturing firms and their capabilities to assist them in being discovered and vetted into the global supply chains\cite{doi:10.1080/10196780500083746}.

\subsection{Challenges in Identifying Manufacturing Service Capabilities}
Manufacturing Service Capability (MSC)~\cite{cao2017demand} is the ability of manufacturing enterprises to effectively integrate and configure various resources, reflecting their proficiency in completing specific tasks. This comprehensive concept spans the entire life-cycle of manufacturing, including design, simulation, and production capabilities. It's evidenced in various forms, from industry-recognized certifications like Capability Maturity Model Integration (CMMI), indicating a commitment to quality and consistency, to specific manufacturing processes such as drilling and milling. MSC also covers the adaptability of manufacturers to serve different industries, like medical industry and automotive industry, and their capacity to work with diverse materials, such as plastics and steel. 

Traditional methodologies constrain the scope of MSC identification, restricting it to a limited scale. Within the business sector, platforms such as Thomasnet~\cite{thomasnet} and Google Maps necessitate that manufacturers independently catalog their competencies. This self-reporting approach slows down the expansion of manufacturing business networks. In the academic context, there is always an assumption that MSC data is pre-defined and uniformly structured for supply-demand matching \cite{cheng2017modeling,cao2017demand}. However, in reality, a significant portion of MSC data, particularly from smaller, local businesses, is derived from their distinct and varied website structures. To effectively address this, there's a critical need to develop a universally adaptable method that can autonomously and efficiently identify MSCs on a much broader scale.

Existing approaches to automatically identifying MSC \cite{belhadi2019understanding} rely on keyword matching or Natural Language Processing (NLP). Keyword matching involves identifying keywords such as ``CNC machining'' or ``injection molding'' on a manufacturer's website. NLP-based methods analyze relevant information from textual data sources, such as websites, catalogs, or documents. For instance, Named Entity Recognition (NER) can be adopted to identify specific MSCs \cite{kumar2022fabner}. The widespread application of various methods in identifying and categorizing businesses and services has greatly contributed to the optimization of supply chains and decision-making processes in the manufacturing domain \cite{SALA2022522,su14031103}.

However, these methods often suffer from two critical limitations: wrong identification and misidentification. Wrong identification occurs when a business or service is incorrectly categorized. For example, a company selling Computer Numerical Control (CNC) machines may be incorrectly identified as a CNC machining provider, even if it cannot provide CNC machining services \cite{okokpujie2019review}. This misclassification can lead to incorrect assumptions about the company's capabilities, resulting in inappropriate decisions and actions by other parties, such as suppliers or potential customers. Misidentification happens when a business or service's capabilities are not fully recognized or understood. For instance,  manufacturers skilled in titanium processing might be integrated into the supply chain for aircraft or medical device production, given that titanium is a frequently used material in the aerospace industry \cite{peters2003titanium} or the medical industry \cite{baltatu2021new}. However, if these capabilities are not accurately identified, the manufacturer may be overlooked for contracts or collaborations in these sectors. It is crucial to uncover, integrate, and utilize hidden information in manufacturing data sources to aid decision-making, and risk management, and gain insights into the flow of goods, materials, and resources through the supply chain.

\subsection{Objectives}
Recently, Knowledge Graphs (KGs) and Graph Neural Networks (GNNs) are of paramount importance in data representation and knowledge extraction \cite{xu2016knowledge,yasunaga2021qa}. KGs effectively manage complex data with interconnected entities, offering scalability and ease of updates. GNNs complement this by adeptly learning from data's complex relationships and patterns, particularly useful in graph-structured data. In the realm of MSC identification, combining KGs and GNNs leads to a more adaptable, precise, and scalable approach. This integration is key in accurately distinguishing the unique characteristics of various manufacturers, thus reducing misidentifications. This synergy has propelled the development of GNN-based systems for digital supply chain representation in manufacturing, as further detailed in recent research \cite{doi:10.1080/00207543.2021.1956697}.

This paper seeks to harness the power of KGs and GNNs to enhance the discoverability of small manufacturing businesses and their MSCs by prospective clients. The identification of MSCs enabled by KG and GNNs significantly aids startups, entrepreneurs, and researchers in gaining insights from manufacturing data to select potential business partners. In this work, we employ an automated approach to construct a Manufacturing Service Knowledge Graph (MSKG) as introduced in \cite{10.1115/MSEC2021-63766}, serving as the foundational framework for our analysis. An MSKG is comprised of two distinct node categories: ``Manufacturer'' and ``Service''. It encapsulates two types of relational links: one that establishes connections between ``Manufacturer'' and ``Service'' nodes, and another that delineates affiliations amongst ``Service'' nodes. The central challenge lies in effectively modeling the problem using a graph-based approach while concurrently enhancing performance through the strategic application of feature engineering methods. The task of modeling the problem is not only selecting the most appropriate architecture but also modifying the setting of models and data especially for addressing the business objective. It also includes designing effective feature representations that can significantly elevate the overall predictive power and generalizability of our method. The main contributions highlighted in this paper are: 

\begin{enumerate}
    \item We introduce a methodology to deduce MSCs by graph-based node classification, offering unique advantages in the realm of graph-based information inference.

    \item We propose a feature engineering approach tailored for MSKGs that enhances the performance of graph-based analysis by aggregating information from nodes' neighborhoods.

    \item We propose to mitigate the issue of node class imbalance in real-world heterogeneous graphs by generating synthetic edges and nodes, which can be generalized to various practical scenarios.
\end{enumerate}

An example to identify MSC using our approach is shown in 
\hyperlink{figure1}{Figure 1}. 
Suppose we aim to determine if a manufacturer possesses the capability to cater to the automotive industry and handle copper processing. Initially, an MSKG is formulated by gleaning textual data from different manufacturing data sources. Then, the graph's nodes and edges are synthesized to balance the node classes of two distinct node classification objectives: ``Does the manufacturer serve the automotive industry?'' and ``Does the manufacturer process copper?''. Following this, we aggregate information from neighboring nodes within the graph. In the final phase, we employ GNN algorithms to train two distinct node classifiers. These classifiers' outcomes then help ascertain the manufacturer's capability in the automotive sector and copper processing. The refinement in the second and third steps ensures enhanced precision in node classification, leading to more accurate insights into manufacturing competencies.

For the rest of the paper, we review the related work in Sec.~\ref{sec:relate}. The studied problem is defined and the details of the proposed method are presented in Sec.~\ref{sec:method}. In Sec.~\ref{sec:exp}, the experiments are conducted to demonstrate the effectiveness of our method. In Sec.~\ref{sec:con}, the limitations and future work of our method are concluded.

\hypertarget{figure1}{}
\begin{figure}[ht]  % 'h' means "here," you can change this to your preferred placement
  \centering
  \includegraphics[width=1\textwidth]{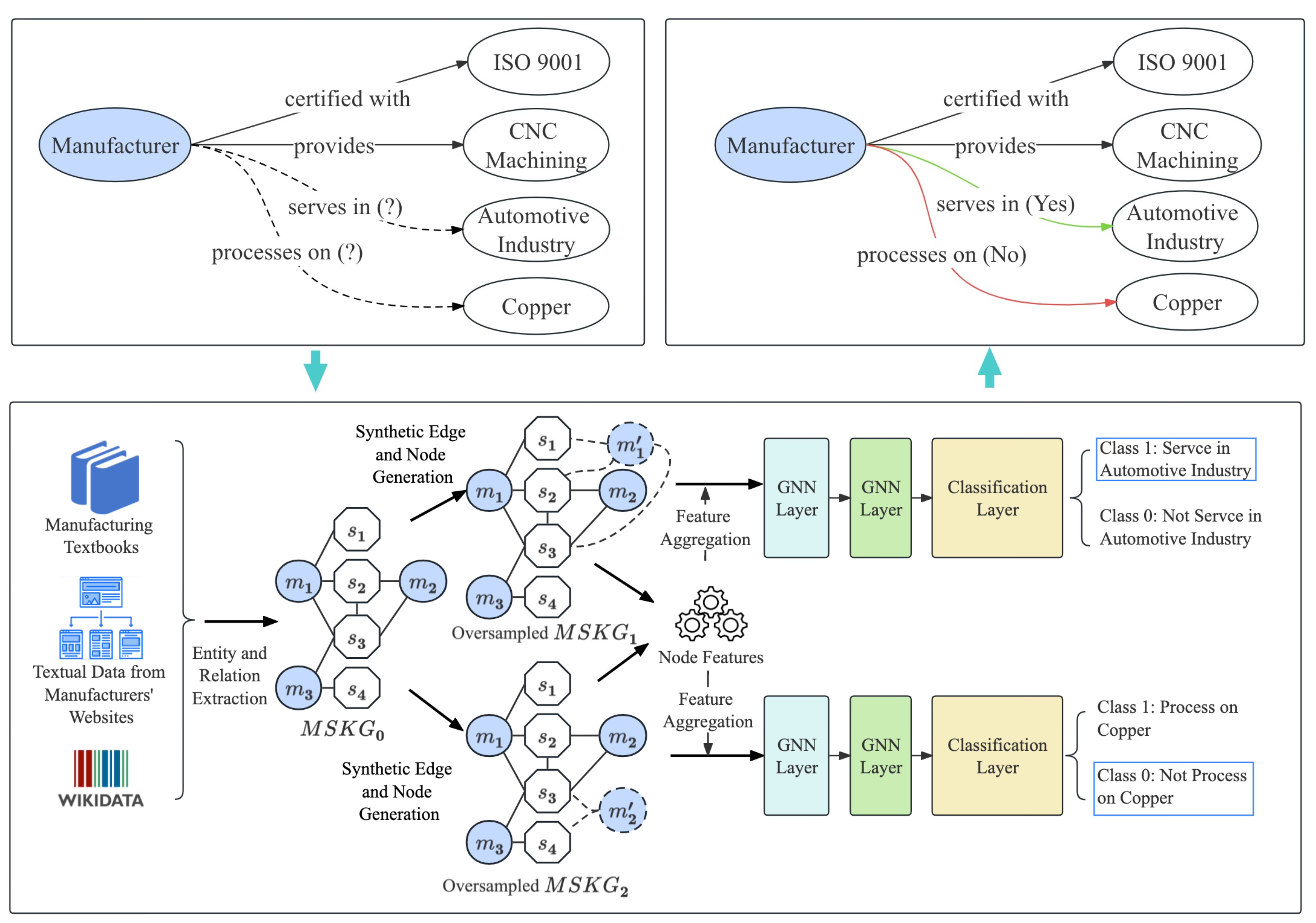} 
  \caption{An Example of Identifying MSC}

  \label{MSCI}
\end{figure}
\section{Related Work}
% \label{sec2}
\label{sec:relate}

%%       \citet{<label>} ==> Jones et al. [21]
%%       \citep{<label>} ==> [21]
%%
\subsection{Manufacturing Data Sourcing and Inference}

Currently, there are various methods for manufacturing data sourcing. Key term matching and Term Frequency-Inverse Document Frequency (TF-IDF)~\cite{ramos2003using} help with extracting essential words from manufacturing textual data~\cite{BOUZARY2020101989}. K-means and Latent Dirichlet Allocation (LDA) algorithms are used for document clustering and topic modeling in manufacturing text data mining~\cite{XIONG2019333}. When handling a large amount of manufacturing textual data, several forms of NLP have been constructed through different kinds of embedding representation schemes. For example, Word2Vec~\cite{goldberg2014word2vec}, Doc2Vec ~\cite{lau2016empirical}, Bidirectional Encoder Representations from Transformers (BERT)~\cite{devlin2019bert}, OpenAI embedding~\cite{muennighoff2022sgpt} are widely used in dealing with content embedding in the manufacturing domain. Named Entity Recognition is applied to identify goods, materials, and resources in different parts of the supply chain~\cite{10.1145/3587716.3587723}.

With the development of the digital supply chain, applications of data-driven inference in the manufacturing domain have been evolving in recent years. There are inference systems constructed which are related to evaluating management of supply chain performance~\cite{pourjavad2018application}, supply chain risks~\cite{carrera2008supply} or downstream demand inference~\cite{tliche2020improved}. Manufacturing information inference is inferring information about a manufacturing process or service based on available data and evidence, which can help suppliers to optimize their cost-effectiveness, and customer satisfaction as well as be noticed and considered by more potential clients. Villas-Boas~\cite{berto2007vertical}conducts inference on vertical relationships between manufacturers and retailers. A framework to enable the reusability of manufacturing knowledge through inference rules applied to manufacturing ontologies is introduced in~\cite{mourtzis2014simulation}. Cao et al.~\cite{cao2017demand} propose a model for estimating MSCs, specifically emphasizing machining and production services. Another framework for canonicalizing MSC models is proposed using the reference ontology ~\cite{kulvatunyou2015framework}. But none of them is about gaining insightful inference on the latent relationships between manufacturers and various manufacturing services.

\subsection{Knowledge Graph Construction from Unstructured Data}
Constructing a KG from unstructured data is more challenging due to the inherent difficulty in accurately extracting entities and relationships from such data. In the healthcare domain, Health KG Builder is introduced by~\cite{zhang2020hkgb}, which can be used to construct disease-specific and extensible health KGs from unstructured sources. Zhu et al.~\cite{zhu2017intelligent} focuses on the application of KG in the traditional geological field and proposes a novel method to construct KG from complex geological unstructured data. Li and Starly~\cite{li2021knowledge} presents a bottom-up approach to parse through unstructured text available on the websites of small manufacturers across the United States to construct a MSKG~\cite{10.1115/MSEC2021-63766}. 

\subsection{Current Graph Neural Networks and Downstream Applications}

GNNs are able to effectively extract complicated, non-linear relationships in datasets. One type of GNN is Message-passing neural networks (MPNNs)~\cite{DBLP:journals/corr/GilmerSRVD17} which utilize a message-passing mechanism to aggregate information from neighboring nodes in the graph and accordingly update the delineation of each node. GraphSAGE~\cite{DBLP:journals/corr/HamiltonYL17} and graph convolutional networks (GCNs)~\cite{zhang2019graph}, SGC-GNNs~\cite{DBLP:journals/corr/abs-1902-07153} and EdgeConv~\cite{DBLP:journals/corr/abs-1801-07829} are MPNNs. APPNP~\cite{DBLP:journals/corr/abs-1810-05997} incorporates personalized PageRank scores into the propagation process to improve prediction accuracy, making it a specialized variant within the MPNNs. Another type is Graph Generative Models (GGM) which are used for generating synthetic graph structures based on probabilistic models such as Markov random fields or Bayesian networks. Instances of GGMs include graph recurrent neural networks (GRNNs)~\cite{Ruiz_2020} and graph generative adversarial networks (GraphGANs)~\cite{wang2018graphgan}. Last is Graph Transformer Models~\cite{yun2019graph} which are built on the transformer framework, such as graph transformers (GTrs)~\cite{DBLP:journals/corr/abs-2111-12696} and graph attention transformers (GATs)~\cite{veličković2018graph}.

With the rapid development of GNNs, they have demonstrated state-of-the-art performance on diverse graph downstream applications. Three typical tasks are node classification~\cite{xiao2022graph}, link prediction~\cite{kumar2020link} and graph classification~\cite{errica2020fair}. Node classification is when we have a KG with a certain ratio of nodes labeled, a classifier is trained on those labeled nodes so that it can classify the unlabeled nodes in the graph. Class imbalance has been an essential challenge in node classification. 

\subsection{Class Imbalance}
Class imbalance has been a vital research topic in machine learning for years. Numerous tasks, like fraud detection~\cite{liu2021pick} and sentiment analysis~\cite{ghosh2019imbalanced} suffer from class imbalance. The main approach is changing the data itself or the way the model is used to solve class imbalance, such as undersampling and oversampling. Undersampling is a technique that reduces the number of instances in the majority class so that it is more evenly represented with the minority class. When removing instances from the majority class, we may lose essential information in those instances which could negatively affect the performance of the models. Oversampling is duplicating existing instances in minority classes which can mitigate class imbalance but result in over-fitting in the data training process. Many variations of oversampling have been proposed to improve its effectiveness. Synthetic Minority Over-sampling Technique (SMOTE)~\cite{chawla2002smote} is one of the most popular over-sampling methods. It involves generating synthetic minority class samples by interpolation to multiply the representation of the minority class. To address node class imbalance issue in a graph, current methods such as GraphSMOTE~\cite{10.1145/3437963.3441720} and GATSMOTE~\cite{math10111799} are proposed, which leverage synthetic data generation to balance class distribution, enabling models to better handle underrepresented classes.

\subsection{Feature Engineering}
Feature engineering is a fundamental component of machine learning that profoundly influences model performance. Data transformation techniques encompass one-hot encoding, feature scaling, etc., which are crucial for managing data heterogeneity and scaling issues. Dimension reduction approaches, such as t-distributed stochastic neighbor embedding (t-SNE)~\cite{van2008visualizing}, Principal component analysis (PCA)~\cite{DBLP:journals/corr/abs-1804-02502} and feature selection, aid in managing high-dimensional data and curating a subset of the most informative features. Additionally, domain-specific knowledge often guides the creation of task-specific features~\cite{tamhane2021feature,e25091278}. The choice of feature engineering method depends on the data type and problem domain. Hence, it is essential to select and customize the strategy of feature engineering based on the characteristics of the KGs in the manufacturing domain to enhance the model's predictive ability in identifying MSCs.

\section{Methodology}
% \label{sec4}
\label{sec:method}

\subsection{Problem Statement}

The objective of our paper is to identify if a manufacturer is capable of a specific potential manufacturing service, which can be converted to a graph-based information inference problem. The reasons are as follows: first, the representations of MSCs is easy to show in graphs. Manufacturers, as well as their services (like machining and automotive industry), can be represented as nodes, while the relationships between them, indicating manufacturers' service capability, can be represented as edges. Second, graph has scalability and flexibility. As  more data about new manufacturers and their services are gathered, the graph can be continuously expanded. For instance, if a new service becomes relevant to the automotive industry, it can be easily incorporated into the graph, allowing for dynamic updating of inferences. Last but not least, the inference can be conducted through connectivity in the graph. If a new manufacturer provides both machining and 3d printing, and these capabilities are commonly found among manufacturers serving the automotive industry (as seen in the graph), then it's likely that this manufacturer also has potential in the automotive domain. This conclusion can be inferred based on the proximity and connectivity patterns in the graph.

In this study, our objective is to infer MSCs of manufacturers, which is achieved by utilizing the connections identified within the MSKG. Our approach offers a detailed assessment of the manufacturers' expertise and proficiency across diverse manufacturing domains. Within the MSKG, the connections between manufacturers and services are classified into four key categories: ``provide'', ``certified with'', ``serve in'', and ``process on''. These categories are directly linked to the targeted manufacturing services, encompassing areas such as manufacturing processes, certifications, industries, and materials. 

\( G = \{M, S, A, F\} \) is the MSKG used for graph-based information inference. The attributed network is depicted by various elements, as follows:

\begin{itemize}
    \item \( M = \{m_1, . . . , m_n\} \) is a set of \( n \) nodes, where each node represents a unique manufacturing business. Within the graph, these nodes can be oversampled, as various manufacturer nodes can potentially link to a group of same manufacturing service nodes. \( M_t \) is a subset of \( M \) used in training.
    
    \item \( S = \{s_1, . . . , s_i\} \) is a set of \( i \) manufacturing service nodes, where each node represents a unique entity so they can’t be oversampled within the graph. \( S_t \) is a subset of \( S \) used in training.
    
    \item \( A \in \mathbb{R}^{p \times p} \) is the adjacency matrix of \( G \), \( p = n + i \). \( A_{km} \) is essentially a binary indicator that tells you whether the vertices corresponding to row \( k \) and column \( m \) are directly connected in \( G \).
    
    \item \( F \in \mathbb{R}^{p \times d} \) is the node attribute matrix, where \( F[j, :] \in \mathbb{R}^{1 \times d} \) is the node attributes of node \( j \). \( d \) is the dimension of the node attributes.
    
    \item \( Y \in \mathbb{R}^p \) represents the class information for nodes in \( G \). \( Y_t \) is a subset of \( Y \) for training. The task of node classification is to predict whether a given node represents a manufacturer as well as it is connected to a designated manufacturing service node. The manufacturer nodes that have direct relationships with these services are labeled as 1, and all other nodes are labeled as 0.
    
    \item \( C = \{c_1, c_2\} \), \( |c_1| \) is the size of majority node class and \( |c_2| \) is the size of minority node class.
    
    \item Imbalance ratio, \( \frac{|c_2|}{|c_1|} \) determines the level of node class imbalance.
\end{itemize}

In the methodology, the inference of MSC is modeled as a GNN-based node classification problem. The proposed method is composed of four subsections: \ref{subsec1} Problem Modeling. It introduces why and how to model our problem into a node classification task over MSKGs. \ref{subsec2} Synthetic Edge and Node Generation. It leverages random sampling and stratified sampling on edges associated with minority classes to balance the node classes to obtain an augmented graph. \ref{subsec3} Feature Aggregation. It aggregates and encodes neighbor nodes’ information through Doc2Vec and t-SNE and combines them with original node features to form the node attribute matrix of the augmented graph.  \ref{subsec4} GNN Classification. It utilizes a GraphSAGE classifier to predict binary class labels of the nodes in the oversampled augmented graph. The following parts provide further details on each step.

%\hypertarget{figure1}{}

%\begin{figure}[ht]  % 'h' means "here," you can change this to your preferred placement
 % \centering
 % \includegraphics[width=1\textwidth]{overall.png}  % Replace 'image_filename' with the actual filename of your image
 % \caption{The Overall Framework of Infering Manufacturing Service Capability}

%\end{figure}

\subsection{Problem Modeling} % Add this line for the subsection
\label{subsec1}    

The process of constructing an MSKG serves as the foundation for our graph-based inferential procedures. The construction of the MSKG is carried out in three phases. First, a web-scraping process is initiated to gather the text content from the manufacturers' websites in the United States to create manufacturer nodes in \( M \). Second, we identify manufacturing service nodes in \( S \) as well as the edges between them such as subclass relationships from Wikidata and standard manufacturing textbooks. Third, after text pre-processing, keyword matching is conducted between \( M \) and \( S \) to obtain the relationships.\( G_0 \) denotes the initial graph constructed. The basic schema structure of an MSKG includes two entity types: manufacturer name, \( M \); 2) manufacturing service, \( S \). \( S \) encompasses the manufacturing process, relevant certifications, materials utilized, and the industries in which the manufacturer operates. Equation (1) is used to initialize the node attribute matrix, designated by $F[j]$: 

\begin{equation}
F[j] = \begin{cases}
    0 & \text{if } j \in M \\
    1 & \text{if } j \in S \text{ and } j \text{ is an industry} \\
    2 & \text{if } j \in S \text{ and } j \text{ is a service} \\
    3 & \text{if } j \in S \text{ and } j \text{ is a material} \\
    4 & \text{if } j \in S \text{ and } j \text{ is a certification}
\end{cases}
\end{equation}

\( G \) is obtained from \( G_0 \) by excluding the target manufacturing service nodes and the edges directly connected to them. This step is essential for node classification. For instance, if we need to infer which manufacturer serves the medical industry, we initially mask the correct answer within the graph. By excluding the ``Medical Industry'' node and its corresponding manufacturer relationships, we can partition the modified graph: some nodes for training a graph-based classifier and others for subsequent prediction and evaluation.

To deduce the MSC from the MSKG, we explore both link prediction and node classification approaches to tackle the challenge. Node classification is selected as the primary method for our study, with link prediction serving as the comparative approach. The reason is that in MSKGs, the number of links is typically 15 times greater than the number of nodes. Given this, computing predictions for every potential link can be computationally intensive. It's essential to save on computational costs, especially for the dynamic nature of MSKGs. To accomplish our primary objective, node classification aims to discern whether a node represents a manufacturer node and is directly linked to a manufacturing service node, while link prediction is designed to predicting the relationships between manufacturer nodes and a designated manufacturing service node. On the other hand, node classification aims to discern whether a node represents a manufacturer and is directly linked to a manufacturing service node.

\subsection{Synthetic Edge and Node Generation} % Add this line for the subsection
\label{subsec2}     % Optionally, add a label for the subsection
Possible ways to oversample the graph could be through node duplication, or adapt SMOTE to the graph data, like GraphSMOTE. These methods are sub-optimal for our graph due to the following reasons: 1) Since each node in \( S \) is unique, directly oversampling entities in \( S \) may distort the overall structure of the MSKG, leading to a misrepresentation of the relationships between nodes. Simply oversampling entities in the graph may easily cause overfitting during the training process as well. 2) Since nodes of \( M \) and \( S \) can be classified into the same node classes, GraphSMOTE or performing SMOTE in the node embedding space may result in the creation of synthetic nodes in the minority class that are neither similar to entities in \( M \) and \( S \). 3) Both SMOTE and GraphSMOTE, regardless of whether synthetic nodes are generated through interpolation in the raw feature space or embedding space, fail to consider the diversity within the same node class in a heterogeneous graph. This implies that within a single node class, there may be different node types, leading to a situation where certain types can be oversampled in while others are not.

\hypertarget{figure2}{}
\begin{figure}[ht]  % 'h' means "here," you can change this to your preferred placement
  \centering
  \includegraphics[width=1\textwidth]{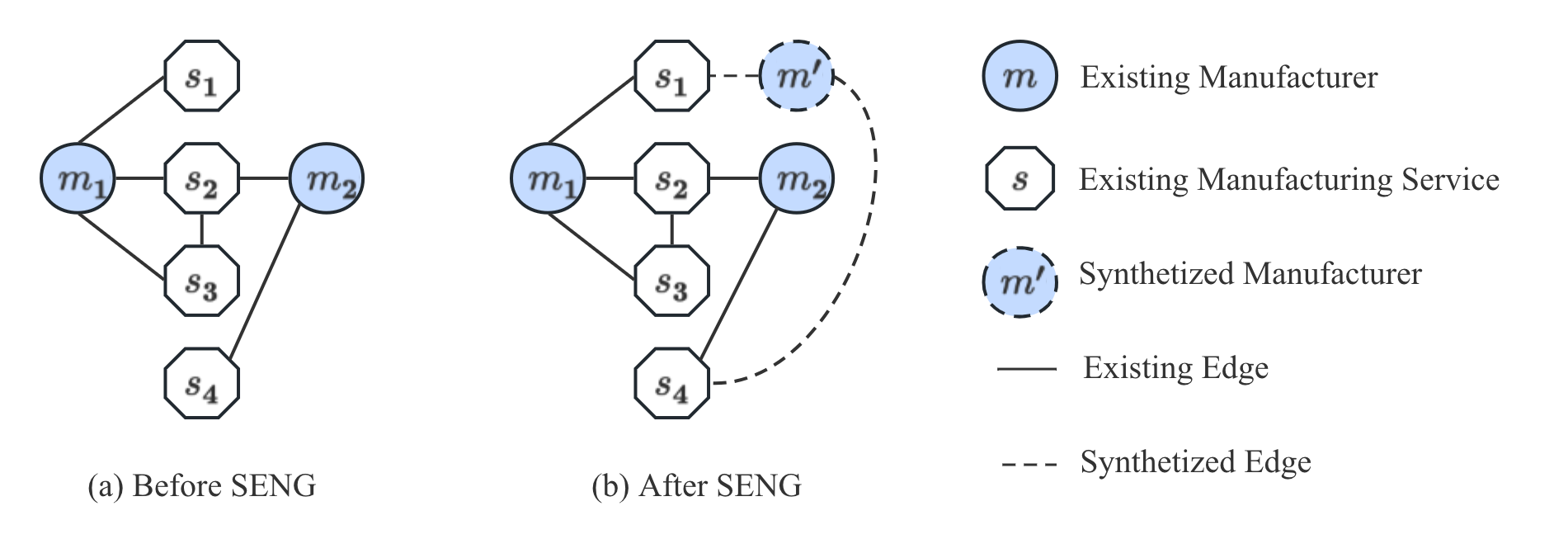}  % Replace 'image_filename' with the actual filename of your image
  \caption{An Example of SENG}
  \label{SENG}
\end{figure}

Hence, we propose a heuristic approach, Synthetic Edge and Node Generation (SENG), to conduct graph oversampling on an MSKG shown in \hyperlink{figure2}{Figure 2}. SENG-oversampling utilizes bagging of entities and edges as a means of mitigating the overfitting that is introduced through oversampling in the training process. To generate a synthetic node along with all the edges associated, there are six steps: 1) To minimize redundancy in the generation of synthetic nodes, a random selection $\alpha$ is made from the set [2, 3, 4]. This choice is deliberately constrained to values greater than 1 and less than 5, thus ensuring a diversified range of selections; 2) Randomly sample $\alpha$ elements from \( M \) with replacement to obtain a subset of \( M \), $M_{\text{sub}}$ ; 3) Use $M_{\text{sub}}$ to get corresponding $A_{\text{sub}}$; 4) Use $A_{\text{sub}}$ to obtain a subset of \( S \),  $S_{\text{sub}}$; 5) Random sample $\frac{1}{\alpha}$ from $S_{\text{sub}}$ ; 6) Create a synthetic node \( m' \) along with synthetic edges connecting \( m' \) with entities in $S_{\text{sub}}$. After this generation process, node classes are more balanced. However only applying SENG may cause severe overfitting during the training process. To improve training performance, we introduce the feature aggregation process in the following step.

\subsection{Feature Aggregation} % Add this line for the subsection
\label{subsec3}    
Word embedding \cite{wang2019evaluating}, a technique based on artificial neural networks, allows textual data to be represented in a way that can be understood by computers, which is achieved by representing words as vectors. Doc2Vec, an extension of the Word2vec model, is a word embedding method that generates a vector representation of a paragraph, allowing for the detection of semantic similarity. This vector representation captures the meaning and context of the words in the document, as well as their order and arrangement. T-SNE is a machine learning algorithm that can project high-dimensional data into a lower-dimensional space as well as preserve the structure of the data. Both Doc2Vec and t-SNE are used in the node feature generation process.

Feature Aggregation (FA) aims to enhance the performance of node classification by enriching node features. Through SENG, \( G \) is transformed into $\widetilde{G}$, with corresponding changes occurring in $\widetilde{M}$, $\widetilde{S}$, $\widetilde{A}$, $\widetilde{F}$, $\widetilde{C}$ and $\widetilde{Y}$. Textual information from the neighboring nodes, which are their names, are collected to populate the representation of manufacturer node features in $\tilde{F}'$. \( D \), a dictionary, which is a built-in data structure that allows the storage and retrieval of key-value pairs. In this paper, \( D \) contains all the entities in $\widetilde{M}$ as keys and their first-order related names of neighbouring nodes in $\widetilde{S}$ as values. Each value, indexed by a key, is a paragraph within a corpus. Each paragraph corresponds to an entity in $\widetilde{M}$. The vectors of paragraph are learned by Doc2Vec such that each paragraph is mapped to a high dimension space, feature matrix $F_1$. Dimensionality reduction via t-SNE is performed to project high dimensional vectors into 2-dimensional space and generate the feature matrix $F_2$. In Equation (2), each row in $F_2$ is integrated with $\widetilde{F} [j]$ where $\widetilde{F}' [ j, : ]\in \mathbb{R}^{1 \times 3}$ is the updated node features of node $j$. 

\begin{equation}
\widetilde{F}'[ j, : ] =
\begin{cases}
    \widetilde{F}[ j ] + F_2[ j, : ], & \text{if } j \in M \\
    \widetilde{F}[ j ] + [0,0], & \text{otherwise}
\end{cases}
\tag{2}
\end{equation}

\subsection{GNN Classification} % Add this line for the subsection
\label{subsec4}    % Optionally, add a label for the subsection
This step is to train a GNN-based node classification model on the augmented graph $\widetilde{G} = \{\widetilde{M}, \widetilde{S}, \widetilde{A}, \widetilde{F}'\}$. The binary labels assigned to each node are determined by assessing whether they establish direct connections with the target MSCs within \( G_0 \). We partition our graph data for training, validation, and testing. Additionally, we employ stratified splitting for the augmented graph data and incorporate it into the training data. The class imbalance problem is addressed in 4.2 through SENG, resulting in equal class support while using a weighted cross-entropy loss in model training. 

GNN-based node classification employs a specialized deep learning framework to categorize or label individual nodes within a given graph. A two-layer GraphSAGE is adopted to derive node embeddings. An output layer is then appended, processing the feature vectors from the final GraphSAGE layer to assign node classification labels. At the first layer, the aggregated embedding \(h_{1}^{N(j)}\) at node \(j\), based on the set of sampled neighbor nodes \(N(j)\), is concatenated with the node's attributes \(\widetilde{F}'[j, :]\) from \(\widetilde{G}\). The equation to generate aggregated information \(h_{1}^{N(j)}\) at node \(j\) is represented as Equation (3). Passing and concatenating the aggregated information with node attributes \(\widetilde{F}'[j, :]\) from \(\widetilde{G}\), a node embedding of \(j\) at the first layer is expressed as Equation (4).

\begin{equation}
h_{N(j)}^1 = \text{MEAN}\left(\{ \widetilde{F}'[u,:] \, \forall u \in N(j), \forall j \in V \}\right)
\tag{3}
\end{equation}

\begin{equation}
h_j^1 = \text{ReLU}\left(W^1 \cdot \text{CONCAT}(\widetilde{F}'[j,:], h_{N(j)}^1)\right)
\tag{4}
\end{equation}

$W_k (k = 1, 2, 3)$ refers to the weight parameters of each layer. The mean aggregator is applied in the aggregated information equation at each layer. Similarly, at the second layer, the aggregated neighbor nodes’ embedding \(h_{N(j)}^2\) at node \(j\) is combined with node \(j\)'s embedding from the previous layer, as depicted in Equations (5) and (6). ReLU is used as the activation function in generating node embeddings at both layers.

\begin{equation}
h_{N(j)}^2 = \text{MEAN}\left(\{ h_u^1, \forall u \in N(j), \forall j \in V \}\right)
\tag{5}
\end{equation}

\begin{equation}
h_j^2 = \text{ReLU}\left(W^2 \cdot \text{CONCAT}(h_j^1, h_{N(j)}^2)\right)
\tag{6}
\end{equation}

In addition, the second layer is appended by a sigmoid layer to predict node labels as expressed in Equation (7). \(P_j\) is the probability that node \(j\) is related to a certain manufacturing service. The classifier is finally optimized by cross-entropy loss as shown in Equation (8). \(V\) is the union of \(\widetilde{M} \cup \widetilde{S}\).

\begin{equation}
P_j^{\prime} = \text{Sigmoid}\left(\text{ReLU}\left(W^3 \cdot \text{CONCAT}(h_j^2, H^2 \cdot \widetilde{A}[:,j])\right)\right)
\tag{7}
\end{equation}

\begin{equation}
L_{\text{node}} = \sum_{j\in V} \left(1(Y_j==1) \cdot \log(P_j)\right)
\tag{8}
\end{equation}

The predicted label of node \(j\), \(Y_j^{\prime}\), is set as:

\begin{equation}
Y_j^{\prime} = 
\begin{cases} 
1, & \text{if } P_j > 0.5 \\
0, & \text{otherwise}
\end{cases}
\tag{9}
\end{equation}

Hence, the objective of our framework is to minimize \(L_{\text{node}}\). \(\emptyset\) is the parameter of the node classifier.

\begin{equation}
\min_{\emptyset} L_{\text{node}}
\tag{10}
\end{equation}

\subsection{Training Algorithm} % Add this line for the subsection
\label{subsec5}     % Optionally, add a label for the subsection
The procedure for executing our framework is outlined in \hyperlink{algo1}{Algorithm 1}. From Line 1 to Line 8, the graph is augmented by SENG. From Line 9 to Line 15, node features are enriched and integrated by FA. From Line 16 to the end, a GraphSAGE classifier is trained on the augmented graph \(\widetilde{G}\). Oversampling Scale (OS) determines how much oversampling is applied to the minority class. For example, if OS is set to 1, it means the number of samples in the minority class is doubled by generating synthetic samples until the class distribution is more balanced. 
\hypertarget{algo1}{}
\begin{algorithm}
\caption{Node Classification on MSKG}
\KwData{G = \{M, S, A, F, C, Y\}}
\KwResult{Predicted node labels Y'}
\If{$\frac{|c_2|}{|c_1|} \leq 0.7$}{
    Number of Oversampling nodes $No = \left(1 + OS\right) \cdot c_2$\;
    \For{$i = 1$ to $No$}{
        $M_i = $ entity set of $\alpha$ random selections from $M$\;
        $S_i = $ node set of entities from $S$ which directly relate to nodes in $M_i$\;
        $S_i' = $ randomly sample $\frac{1}{\alpha}$  of the elements from $S_i$\;
        $S_i' = \text{set}(S_i')$\;
        Connect synthetic node $i$ to elements in $S_i'$, update $G$ to augmented \(\widetilde{G}\);
    }
}
\For{node $q$ in $M'$}{
    $S_q = $ node set of entities from $S'$ which directly relate to node $q$\;
    $D[q].\text{append}(S_q)$\;
}
Use $D$ to train Doc2Vec, obtain $F_1$\;
$F_2 \leftarrow \text{t-SNE}(F_1)$\;
\For{node $j \in \widetilde{M} \cup \widetilde{S}$}{
    Generate $\widetilde{F}' [ j, : ]\in \mathbb{R}^{1 \times 3}$ based on Equation (2)\;
}
Randomly initialize $W_k$\;
\While{Not Converged}{
    Learn node embeddings according to Equation (3) - (6)\;
    Update the model using $L_{\text{node}}$\;
}
Return trained node classifier\;
\end{algorithm}

The design of our algorithm has four advantages:  1)The primary advantage is its dynamic adaptability, facilitated by the automated updating of MSKG combined with the utilization of GraphSAGE for inductive learning, which allows for the continual integration of evolving manufacturing service capabilities. 2) It is simple to implement SENG over a minority class on a heterogenous graph or a bipartite graph without distortion appearing during the oversampling process. 3) The utilization of FA significantly enhances the representation of nodes within the graph and subsequently improves the performance of node classification. 4) SENG and FA are applied independently, and either can be removed from the whole algorithm if necessary.

\section{Experiments}
% \label{sec5}
\label{sec:exp}

In this section, we conduct experiments to assess the effectiveness of the proposed method for inferring relationships between manufacturers and manufacturing services. In the experimental evaluation, both real-world datasets and the datasets augmented from real-world datasets with imbalanced class distributions are utilized. Specifically, the following questions are addressed in this study:
\begin{enumerate}\label{Q}
    \item How does the node classification result in the performance of MSC identification compared with link prediction?
    \item Is our method pervasive to a different classifier structure?
    \item How does the utilization of FA in our method result in performance compared with other feature engineering approaches?
    \item How does the performance of our method vary under different OSs in the imbalanced node classification task?
    \item Is our method pervasive to different imbalance ratios?
\end{enumerate}
The experimental settings, including datasets, baselines, configurations and evaluation metrics are presented in \ref{5.1}. Question (1)-(5) are addressed in \ref{5.2} - \ref{5.6} respectively.

\subsection{Settings} % Add this line for the subsection
\label{5.1}  

\subsubsection{Datasets}
We conduct experiments based on a MSKG to identify if manufacturers are capable of the following manufacturing services: ``Machining'', ``Copper'', ``Heat Treatment'' and ``ISO 9001''. The MSKG \cite{Li2023}, containing 7,052 nodes and 112,873 relationships, has been constructed by keyword matching between textual data from over 7,000 manufacturers' websites in the United States as well as common manufacturing services, which are selected from Wikidata and the manufacturing textbooks. 

The task of node classification within the MSKG is to predict whether a given node represents a manufacturer as well as it is connected to a certain manufacturing service or not. The manufacturer nodes that have direct relationships with these services are labeled as 1, and all other nodes are labeled as 0. Once this labeling has been performed, the direct relationships between the manufacturer nodes and the selected manufacturing services are removed from the graph. Node class distributions of the datasets regarding selected manufacturing services are shown in \hyperlink{table1}{Table 1}. Classes in these datasets follow a genuine imbalanced distribution. For each dataset, we split graph data for training, validation and testing following an 8:1:1 ratio. In \ref{5.5} and \ref{5.6}, the datasets generated from original datasets are varied by changing the oversampling ratio and imbalance ratio to analyze the performance of the proposed method under different imbalanced scenarios. 

\hypertarget{table1}{}
\begin{table}[htbp]
\centering
\caption{Node Class Distributions of the Datasets}
\footnotesize  % Use the footnotesize font command for a smaller font
\begin{tabular}{>{\centering\arraybackslash}p{3cm} >{\centering\arraybackslash}p{3cm} >{\centering\arraybackslash}p{3cm}}
\toprule
\textbf{Datasets} & \textbf{Majority Class} & \textbf{Imbalance Ratio (\%)} \\
\midrule
Machining & 1 & 58.66 \\
Copper & 0 & 19.00 \\
Heat Treatment & 0 & 27.30 \\
\bottomrule
\end{tabular}
\end{table}

The link prediction on the MSKG is to predict whether an edge between a manufacturer node and a designated service node exists or not. The edges between the manufacturer nodes and the selected manufacturing services are split following an 8:1:1 ratio for training, validation and testing. The rest of the edges in the graph are added to the training data.

\subsubsection{Baselines}

The performance of our proposed method is evaluated in comparison to alternative solutions for identifying MSCs on the MSKG. These solutions can be used to establish a benchmark for highlighting the improvement or added value of our work. For link prediction tasks, not only GraphSAGE, GCN, EdgeConv, SGC, and APPNP are used as GNN classifiers and compared, but also FA component is utilized on GraphSAGE and GCN to see if integrating FA can enhance the performance of link prediction. For node classification tasks, we assess the performance of a GraphSAGE Classifier in comparison to the following methods:

\begin{itemize}
    
    \item \textbf{$GraphSAGE$}: A GraphSAGE node classifier trained and tested on an imbalanced dataset without any pre-processing or balancing techniques applied.
    
    \item \textbf{$SENG-GraphSAGE$}: Utilizes the SENG part of our method by generating synthetic nodes and edges in the node classification training process to mitigate class imbalance issues but excludes the FA component.
    
    \item \textbf{$FA-GraphSAGE$}: Utilizes the FA component of our method to improve the performance of node class classification on an imbalanced dataset by enriching node features.
    
    \item \textbf{$SF-GraphSAGE$}: Utilizes both SENG and FA components of our method to improve the performance of node class classification on an imbalanced dataset.
\end{itemize}

\subsubsection{Configurations}

All experimental trials were conducted within a consistent Google Colab environment, employing the ADAM optimization algorithm~\cite{kingma2014adam} for training the models. The learning rate for all models was set to 0.01. All models were trained until convergence, with the maximum training epoch set to 415. The OS was fixed at 1 for all datasets in \ref{5.2}.

\subsubsection{Evaluation Metrics}

We adopt two criteria for evaluating imbalanced classification, in line with previous studies: Area Under the Receiver Operating Characteristic curve (AUC-ROC) and Area Under the Precision-Recall curve (AUC-PR). AUC-ROC metric measures the ability of a classifier to distinguish between the positive and negative classes by comparing the true positive rate and false positive rate. AUC-PR illustrates a model's ability to distinguish between positive and negative classes by comparing precision and recall.

\subsection{Overall Performance of MSC Identification} % Add this line for the subsection
\label{5.2}
The investigation of the overall performance of MSC predictions is conducted to answer Question (1). To mitigate the effects of randomness, each experiment is repeated on multiple occasions, with a minimum of three iterations. For link prediction tasks, according to \hyperlink{table2}{Table 2}, GCN consistently performs well across most evaluation metrics and datasets, which indicates that GCNs are a robust choice for link prediction tasks, as they can capture complex relationships in the graph structures effectively. FA-GraphSAGE and FA-GCN generally outperform their non-feature-aggregated counterparts. This suggests that incorporating additional features into the graph-based models can lead to improvements in link prediction. For node classification tasks, according to \hyperlink{table3}{Table 3}, the combination of both SENG and FA components in SF-GraphSAGE leads to the highest AUC-ROC and AUC-PR scores in most cases. This demonstrates that utilizing both SENG and FA can significantly improve performance on the imbalanced datasets. Besides, it is noticed that utilizing SENG without incorporating FA may not yield an improvement in performance due to the insufficiency of node features. On the contrary, using FA independently can greatly improve evaluation results from the baseline by augmenting node features. Additionally, extra evaluations are conducted across varying Train-Test-Valid Ratios (8:1:1, 7:1.5:1.5, 6:2:2, 5:2.5:2.5) using two node classification models: GraphSAGE and FA-GraphSAGE. It is consistently observed that FA-GraphSAGE outperformed GraphSAGE across all metrics for each Train-Test-Valid Ratio. In summary, these results underscore the importance of our method in the context of inferring MSCs, particularly when integrated with node classification. As a result, in the subsequent experiments, we will concentrate on analyzing the variations in node classification.

\hypertarget{table2}{}
\begin{table}[htbp]
\centering
\caption{Comparison of Different Methods for Link Prediction}
\begin{adjustbox}{width=\textwidth}
\small
\begin{tabular}{lcccccc}
\toprule
Datasets & \multicolumn{2}{c}{Machining} & \multicolumn{2}{c}{Copper} & \multicolumn{2}{c}{Heat Treatment} \\
\cmidrule(lr){2-3} \cmidrule(lr){4-5} \cmidrule(lr){6-7}
Evaluation Metrics & AUC-ROC(\%) & AUC-PR(\%) & AUC-ROC(\%) & AUC-PR(\%) & AUC-ROC(\%) & AUC-PR(\%) \\
\midrule
GraphSAGE & 19.77 & 30.64 & 37.05 & 31.11 & 17.30 & 27.86 \\
GCN & 27.82 & 42.06 & 39.98 & 40.84 & 44.08 & 42.31 \\
EdgeConv & 34.67 & 33.01 & 39.78 & 31.94 & 20.06 & 28.20 \\
SGC & 22.83 & 31.10 & 35.18 & 31.25 & 34.14 & 30.69 \\
APPNP & 17.44 & 39.26 & 11.83 & 32.96 & 10.06 & 32.26 \\
FA-GraphSAGE & 30.40 & 43.00 & 44.34 & 43.12 & 40.95 & 42.12 \\
FA-GCN & \textbf{37.95} & \textbf{45.45} & \textbf{49.13} & \textbf{45.20} & \textbf{54.10} & \textbf{47.02} \\
\bottomrule
\end{tabular}
\end{adjustbox}
\end{table}

\hypertarget{table3}{}
\begin{table}[htbp]
\centering
\caption{Comparison of Different Methods for Node Classification}
\begin{adjustbox}{width=\textwidth}
\small
\begin{tabular}{lcccccc}
\toprule
Datasets & \multicolumn{2}{c}{Machining} & \multicolumn{2}{c}{Copper} & \multicolumn{2}{c}{Heat Treatment} \\
\cmidrule(lr){2-3} \cmidrule(lr){4-5} \cmidrule(lr){6-7}
Evaluation Metrics & AUC-ROC(\%) & AUC-PR(\%) & AUC-ROC(\%) & AUC-PR(\%) & AUC-ROC(\%) & AUC-PR(\%) \\
\midrule
GraphSAGE & 61.10 & 55.40 & 51.60 & 16.13 & 52.97 & 22.92 \\
SENG-GraphSAGE & 53.90 & 50.10 & 54.74 & 17.23 & 56.42 & 38.72 \\
FA-GraphSAGE & 78.70 & 71.80 & 73.56 & \textbf{43.48} & 78.47 & 52.51 \\
SF-GraphSAGE & \textbf{82.80} & \textbf{80.90} & \textbf{76.42} & 41.92 & \textbf{79.78} & \textbf{53.25} \\
\bottomrule
\end{tabular}
\end{adjustbox}
\end{table}

\subsection{Influence of Classifier}
To answer Question (2), we undertake an analysis of the performance variations of the models replacing GraphSAGE with GAT and GCN. All experiments have the same configuration settings as GraphSAGE, are implemented on the same GAT and GCN. ``Machining'' is used as the target manufacturing service, with its original class distributions and OS set to 1. The results presented in \hyperlink{table4}{Table 4} reveal that both FA-FCN and SF-GCN, have good performance in evaluation metrics for identifying the capability of ``Machining''. The difference in their mechanisms for aggregating information from neighbors in the graph could impact their performance in identifying MSC. FA-GCN also outperforms FA- GraphSAGE. The difference in their mechanisms for aggregating information from neighbors in the graph could impact their performance in identifying manufacturing service capability. GAT has a lower performance in conjunction with SENG and FA compared to GCN and GraphSAGE. The potential reason is that it uses an attention mechanism to weigh and aggregate information from neighboring nodes in the graph. The attention mechanism's performance depends on the specific dataset and whether the relationships between nodes benefit from such fine-grained weighting. It might not perform as well as the simpler neighborhood aggregation used in GCN. 
\hypertarget{table4}{}
\begin{table}[htbp]
\centering
\caption{Comparison of Different Classifiers for Node Classification}
\label{tab:comparison}
\small
\begin{tabular}{lcc}
\toprule
\multicolumn{1}{c}{Methods} & \multicolumn{1}{c}{AUC-ROC (\%)} & \multicolumn{1}{c}{AUC-PR (\%)} \\
\midrule
GAT & 56.76 & 55.96 \\
SENG-GAT & 56.49 & 54.98 \\
FA-GAT & 73.81 & 73.98 \\
SF-GAT & 73.11 & 72.96 \\
GCN & 72.48 & 63.99 \\
SENG-GCN & 61.77 & 65.19 \\
FA-GCN & 84.59 & 75.07 \\
SF-GCN & \textbf{85.79} & \textbf{85.96} \\
\bottomrule
\end{tabular}
\end{table}

\subsection{Influence of FA}
To answer Question (3), firstly, we compare the performance of utilizing FA and traditional node feature engineering, which is to convert the names of nodes to node features. All the experiments are conducted with Doc2Vec and t-SNE except the ways of generating node features are different. GraphSAGE and GAT are selected as the classifiers. When applying 
node feature engineering on detecting the capability of ``Machining'' with GraphSAGE, AUC-ROC and AUC-PR are 57.48\% and 62.12\%, respectively. Applying 
the same method with GAT, AUC-ROC and AUC-PR are 56.76\% and 55.96\%. It is noticed that using the FA method significantly outperforms the traditional method in terms of both AUC-ROC and AUC-PR. FA aggregates neighbor service nodes' names to manufacturer nodes, which takes into account the broader context of each manufacturer node, while the traditional approach considers each node's name as its sole feature, more isolated in its context. The results highlight the importance of considering the broader context in graph-based node classification tasks. 

In addition, we undertake an analysis of the performance variations of the algorithms replacing Doc2Vec with Bert. The evaluation metrics obtained from the experiments are reported in \hyperlink{table5}{Table 5}. All experiments have the same configuration settings except Doc2Vec is replaced by Bert in FA. ``Heat Treatment'' and ``Copper'' are selected as target manufacturing services, with the datasets' original imbalance ratios and the OS set to 1. The results presented in Table 4 reveal that the implementation of Doc2Vec in FA exhibits a more substantial enhancement relative to the baseline in evaluation metrics when compared to the utilization of Bert. The utilization of Bert demonstrates a relatively inferior performance on the AUC-PR metric compared to the application of Doc2Vec. Given that the textual data utilized in FA does not consist of complete paragraphs with context-dependent sentences, the requirement for contextual understanding of words within a sentence is diminished, resulting in Bert being less appropriate for the task in comparison to Doc2Vec. 
\hypertarget{table5}{}
\begin{table}[ht]
\centering
\caption{Comparison of Different Solutions to Node Classification with Bert}
\label{tab:node-classification}
\resizebox{\textwidth}{!}{%
\begin{tabular}{@{}lcccc@{}}
\toprule
Datasets & \multicolumn{2}{c}{Copper} & \multicolumn{2}{c}{Heat Treatment} \\
\cmidrule(lr){2-3} \cmidrule(lr){4-5}
Methods & AUC-ROC (\%) & AUC-PR (\%) & AUC-ROC (\%) & AUC-PR (\%) \\
\midrule
FA-GraphSAGE & 63.99 & 22.61 & 54.22 & 23.57 \\
SF-GraphSAGE & 77.93 & 34.61 & 77.65 & 47.01 \\
\bottomrule
\end{tabular}%
}
\end{table}

\subsection{Influence of Oversampling Ratio}
\label{5.5}
In this section, we undertake an analysis of the performance variations of two algorithms which include an oversampling process with respect to varying levels of oversampling, to address Question (4). The oversampling scale is manipulated to take on the values of  \( \{0.2, 0.4, 0.6, 0.8, 1.0, 1.2\} \). For all the experiments in this section, the dataset of ``Machining'' is used with its original imbalance ratio of 0.5866. In order to ensure statistical validity, each experiment was repeated more than 3 times, with the average results presented in \hyperlink{figure3}{Figure 3}. For SENG-GraphSAGE, as the OS is smaller than 1, the evaluation metrics decrease slightly first, then increase and achieve optimal scores at 1. It indicates the improvement in the model's ability to distinguish between Class 1 and Class 0 when synthesizing more samples for minority classes with an OS between 0.6 and 1, as well as shows the fluctuations in the model due to graph oversampling without applying content embedding. When the OS exceeds 1, a degradation in the performance of SENG-GraphSAGE is observed. This phenomenon can be attributed to an excessive generation of synthetic nodes with redundant or similar information, which ultimately hinders the ability of GraphSAGE to learn effectively. For SF-GraphSAGE, as the oversampling ratio increases from 0.2 to 0.8, the evaluation metrics increase, indicating an improvement in the model's ability to distinguish between Class 1 and Class 0. However, as the oversampling ratio increases from 1 to 1.2, the evaluation metrics decrease since too many synthetic nodes are generated with similar attributes, which ultimately impairs the ability of GraphSAGE to learn effectively.
\hypertarget{figure3}{}
\begin{figure}[ht]  % 'h' means "here," you can change this to your preferred placement
  \centering
  \includegraphics[width=1\textwidth]{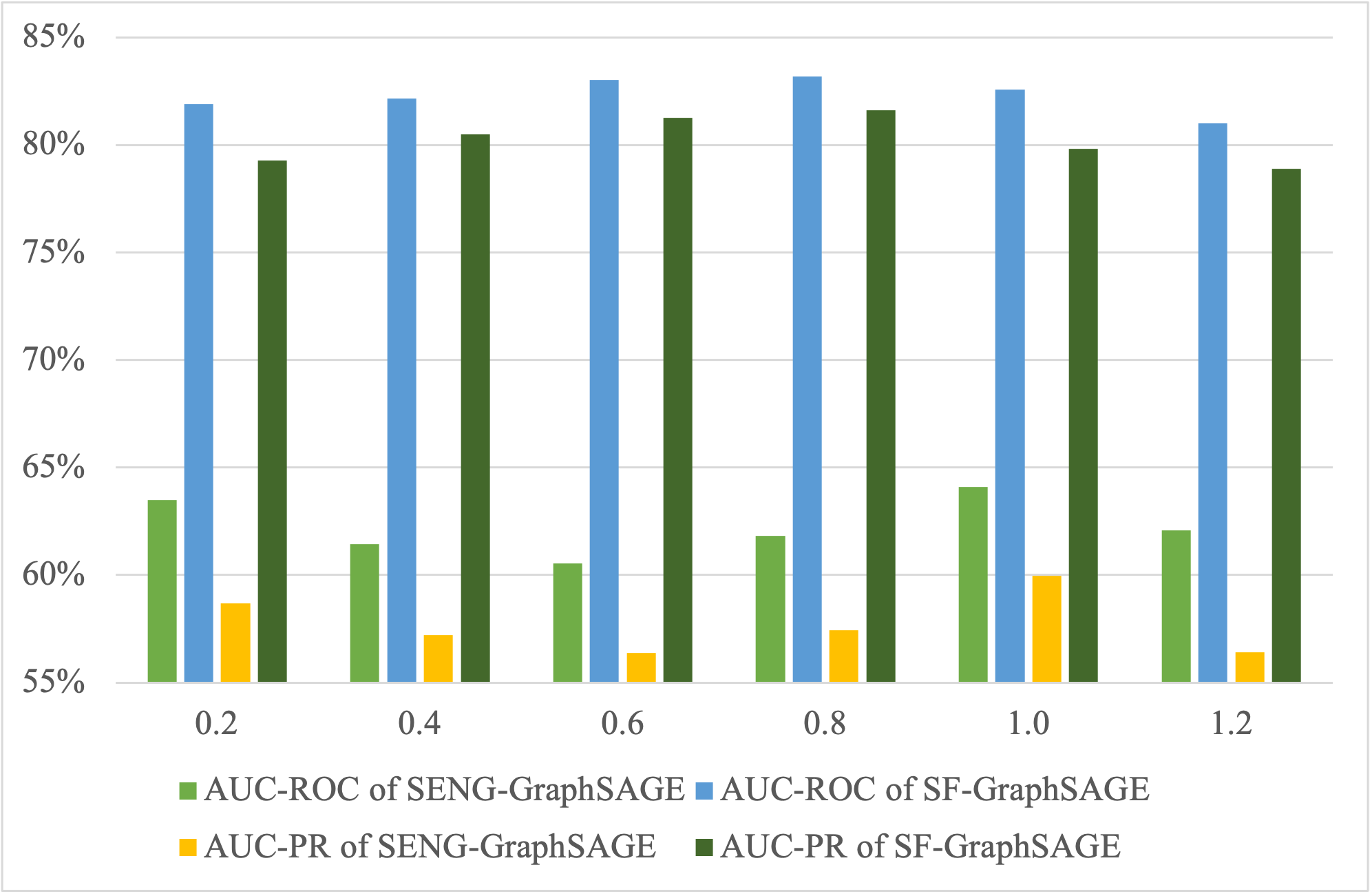}  % Replace 'image_filename' with the actual filename of your image
  \caption{Evaluation Metrics under Different OSs}
  \label{OS}
\end{figure}

\subsection{Influence of Imbalance Ratio}
\label{5.6}
In this section, we undertake an analysis of the performance variations of various algorithms with respect to varying levels of imbalance ratio, to address Question (5). For all the experiments in the section, the ``Machining'' dataset is used as well as a fixed OS of 1 is applied to SENG-GraphSAGE and SF-GraphSAGE. The imbalance ratio scale is manipulated to take on the values of \( \{{0.1, 0.2, 0.4, 0.5866}\} \), where 0.5866 is the original imbalance ratio of the "Machining" dataset. In order to ensure statistical validity, each experiment was repeated more than 3 times, with the average results presented in \hyperlink{figure4}{Figure 4}.

\hypertarget{figure4}{}
\begin{figure}[ht]  % 'h' means "here," you can change this to your preferred placement
  \centering
  \includegraphics[width=1\textwidth]{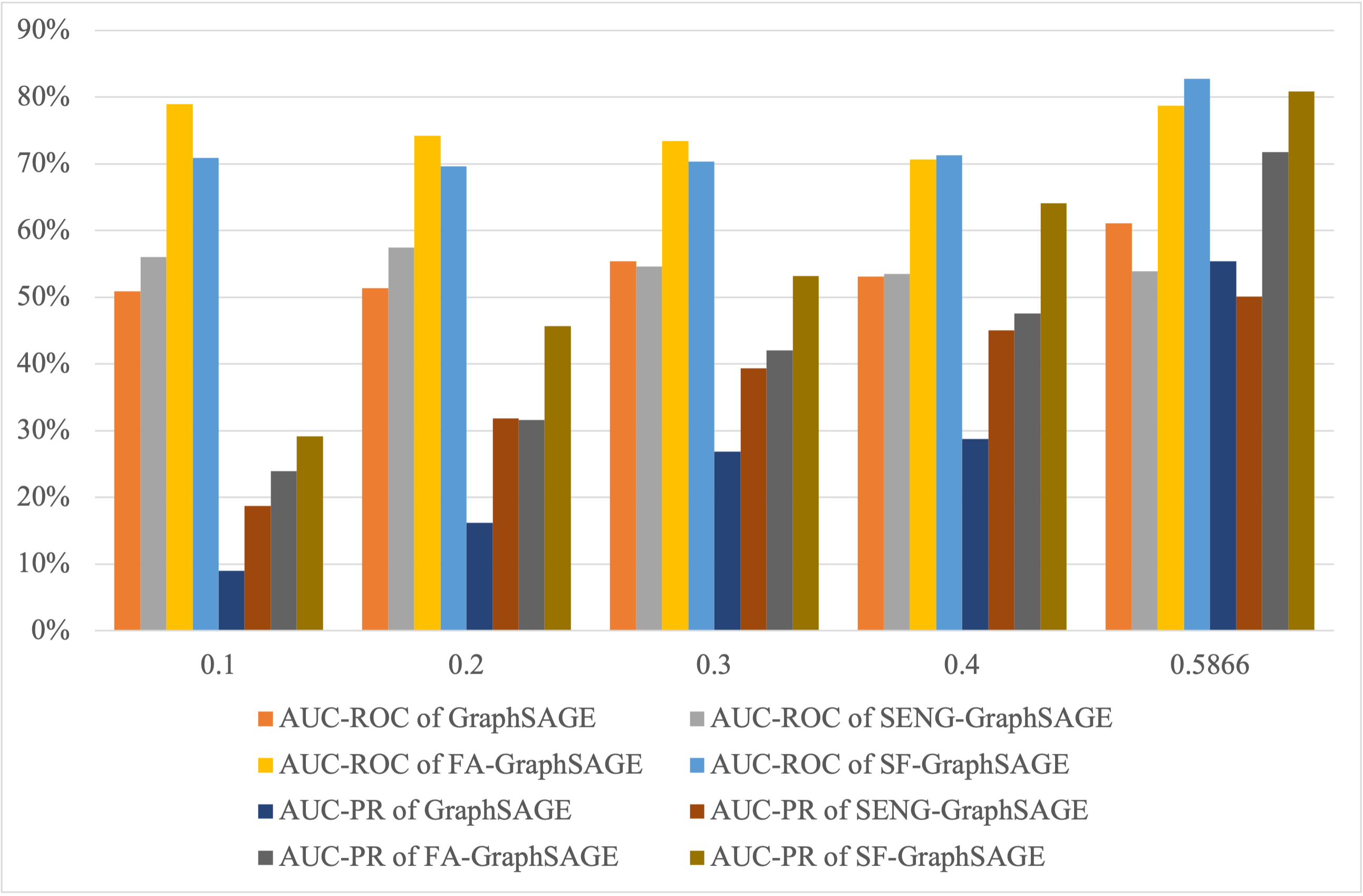}  % Replace 'image_filename' with the actual filename of your image
  \caption{Evaluation Metrics under Different Imbalance Ratios}
  \label{OSR}
\end{figure}

It is noticed that the SF-GraphSAGE method demonstrates a pervasive performance across various imbalance ratios. It not only excels at specific ratios but maintains a competitive edge throughout the range of presented ratios. SF-GraphSAGE exhibits a relatively stable and consistent behavior, especially highlighted in the AUC-PR progression. For FA-GraphSAGE, in terms of AUC-ROC and AUC-PR, FA-GraphSAGE showcases strong performance, especially at the extremes of the given imbalance ratio (0.1 and 0.2). There's a noticeable dip in performance as we move from an imbalance ratio of 0.1 to 0.4, but the method shows adaptability to varying degrees of imbalance, evidenced by its consistently high AUC-ROC and increasing AUC-PR values.

\section{Conclusion And Future Work}

\label{sec:con}

In the realm of industrial engineering and logistics, understanding a manufacturer's service capabilities is crucial for optimizing production efficiency and supply chain management. The current prevailing methods for identifying MSCs from manufacturers are predominantly based on keyword matching and semantic matching. However, these methods tend to either lose hidden information or misunderstand the information, which subsequently leads to incomplete identification of manufacturers' capabilities. To mitigate the limitation,
this study presents a novel GNN-based approach for effectively identifying MSCs within KGs. To enhance the accuracy and performance of this identification process, an innovative strategy is introduced, which involves aggregating information from neighboring nodes and oversampling the graph data. Our rigorous evaluations, conducted on MSKGs, along with subsequent ablation studies, provide unequivocal evidence of the effectiveness and robustness of our proposed approach. These advancements are applicable to a wide range of recommender systems.

Although the effectiveness of our method is demonstrated, some limitations and implications need further attention. In future work, it would be valuable to utilize manufacturing ontologies \cite{lemaignan2006mason} for constructing a more comprehensive MSKG that includes other critical entities like accuracy requirements and material specialization. This enhanced approach in evaluating MSC will ensure that manufacturers are selected not just for their ability to provide manufacturing services, but also for their alignment with the specific and varied needs of different projects. Besides, the study mainly considers a single type of heterogeneous graph. It is imperative to broaden the scope of our method to encompass other heterogeneous graphs or bipartite graphs. Integrating our method with heterogeneous GNN models \cite{10.1145/3292500.3330961} or bipartite GNN models \cite{li2020hierarchical} can be developed. This work simplifies MSKGs and it does not take into account the diversity and directionality of relationships within the graph. The consideration of the difference between edges and the directionality of edges may lead to a more optimized representation of nodes and edges~\cite{DBLP:journals/corr/abs-1904-08745}, which can benefit graph-based downstream tasks. Not only node classification but also other graph-based downstream tasks, such as link prediction and graph classification, are suffering from imbalance class issues. Our methods can be tailored to address other imbalanced class problems, enhancing its efficacy in accurately discerning MSCs.

%\appendix

\bibliographystyle{unsrtnat} % APA-style bibliography
\bibliography{cas-refs} % Use the name of your .bib file





% \end{thebibliography}
\end{document}